\documentclass[a4paper,10pt]{article}

\usepackage[english]{babel}
\usepackage[utf8]{inputenc}
\usepackage{amsmath}
\usepackage{graphicx}
\usepackage{hyperref}

\title{TimeGym: Debugging for Time Series Modeling in Python}
\author{Diogo Seca}
\date{January 13, 2021}
\begin{document}
\maketitle

\begin{abstract}
We introduce the TimeGym Forecasting Debugging Toolkit, a Python library for testing
and debugging time series forecasting pipelines. 
TimeGym simplifies the testing forecasting pipeline by providing generic tests for forecasting pipelines fresh out of the box. These tests are based on common modeling challenges of time series. 
Our library enables forecasters to apply a Test-Driven Development approach to forecast modeling, using specified oracles to generate artificial data with noise.

\vspace{1em}
\end{abstract}

\section{Introduction}

    Test-Driven Development is a software development process in which unitary tests are written before the application code. The goal then becomes to develop the necessary code to pass these tests. The most common scenario is where the developer knows which output the function should return for a given input, i.e. the test oracle is specified. However, applying the same software development process to forecasting or predictive machine learning is difficult. This is because test oracles are oftentimes only partially observable due to issues such as noise and missing values.
    
    
    \subsection{Forecasting and Machine Learning}
    Machine learning methods have recently risen in popularity for tasks of forecasting time series. XGBoost, a decision trees ensemble learning method won the M5 Forecasting Competition, out-performing statistical methods such as Auto-ARIMA, and Auto-ETS \cite{makridakis2020m5}.
    Forecasting competitions such as the M4, M5, or the Global Energy Forecasting Competition 2014 provide an interesting way to benchmark different approaches to the same problems \cite{Hong2016ProbabilisticEF} \cite{HYNDMAN20207}. The forecasting approaches are ranked according to their performance on the dataset. However, there is some doubt whether the results of the best-ranked approaches are not dataset-specific \cite{BONTEMPI2020201}. Bontempi G. poses the question: "Is the outcome of a competition more informative about the quality of the competing algorithms or about the nature of the data that are used for the benchmark?". That is if we were to change the selection of datasets for these competitions, would the best-ranked approaches remain relevant? 
    
    Interestingly, methods that are based on a massive averaging of simple forecasting techniques (e.g. exponential smoothing) have resulted in outstanding results, despite their simplicity when compared to machine learning or deep learning alternatives. This begs the question: what forecasting problems are machine learning methods best suited for?
    
    
    \subsection{Testing Machine Learning}
    
    Durelli et al. found that ML algorithms have recently been used Machine Learning has been recently used for test-case generation, refinement, and evaluation \cite{Durelli2019MachineLA}. And vice versa, software engineering has also contributed to improving AI/ML workflows \cite{Amershi2019-dl}. Braiek et al. and Zhang et al. adapted concepts from the software testing domain to help the machine learning community detect and correct faults in ML programs, namely code coverage, mutation testing, and metamorphic testing \cite{Braiek2020-wf} and \cite{Zhang2020-ns}.
    
    Metamorphic testing offers a way to detect modeling issues during models, by transforming the data. Xie et al. formulated 11 generic metamorphic relations, covering users’ generally expected characteristics that should be possessed by machine learning systems \cite{Xie2020METTLEAM}. Then, they tested the models to unveil the (possibly latent) characteristics of various machine learning systems.
    
    
    With the recent growth in machine learning research, forecasting methods and tools also improved, with more developments expected to come in the future \cite{Petropoulos2020ForecastingTA}. Amershi et al. suggest focusing not only on debugging but also on error analysis \cite{Amershi2019-dl}.
    
    \subsection{Synthetic data}

    Humbatova et al. suggest introducing perturbations in either real or synthetic data, to test the behavior of a learning algorithm \cite{Humbatova2020-pe}. 
    
    However, when using real data to optimize a pipeline, practitioners run into the frequent problem of multiple hypothesis testing. The more tests are performed on the data, and used to optimize the performance over the same data, the higher the risk of overfitting and failing to generalize out-of-sample. This phenomenon has led to the development of the Bonferroni correction method, which consists in adjusts the statistical significance level to consider the number of tests run \cite{weisstein2004bonferroni}.
    
    In the case of synthetic time series, the data can change every test. We need only generate different time series with the same underlying logic. Because we have an infinite amount of time series to sample from, the Bonferroni correction is not required, unlike in real-world datasets in which overfitting is a frequent problem \cite{bailey2014pseudo}.
    
    Moreover, because we can specify the oracle, we can know what the optimal forecast is - which is impossible when using real datasets due to oracle issues such as noisy datasets \cite{seca2021}.
    
    \subsection{Contribution}
    
    No automated testing or debugging library existed for forecasting pipelines. \textit{TimeGym} is an open-source library for debugging forecasting pipelines. \textit{TimeGym} fully abstracts the generation of synthetic time series and the evaluation of forecasts. Moreover, \textit{TimeGym} provides the explainability to users through plots giving users the visual feedback on where forecasting was most off-target. Our library allows forecasters to start applying test-driven development to forecast modeling with a few lines of code.
    
    \subsection{Paper structure}
    
    Section 2 describes the method for generating automated tests. Section 3 demonstrates automated testing on two forecasting pipelines and compares their pros and cons. Section 4 concludes on the impact of this research and presents suggestions for practitioners and researchers.


\section{Automated tests for forecasting pipelines}

    
    We introduce \textit{TimeGym}, a python package for testing and debugging forecasting methods. This package provides tools for stress-test the performance of forecasting pipelines when faced with common forecasting patterns. These patterns are often segregated into components.

    Pipeline are tested several times on different challenges. Each challenge generates three time series to be split for training and test.
    
    The test suites start by generating simple time series with few components. If a forecasting method fails to recognize the Trend, then there's no point in testing whether the method recognizes Trend and Seasonality.
    
    \subsection{Challenges and components}
    
    TimeGym randomly generates time series with known challenges \cite{wheelwright1998forecasting} \cite{hyndman2018forecasting}. These challenges include reusable time series components: 
    
    \begin{enumerate}
        \item gaussian or uniform noise
        \item trend
        \item seasonality
        \item cyclicality
        \item concept drift \cite{gama2014survey}
        \item bimodal or multimodal distributed data
        \item missing values
        \item anomalies / outliers
    \end{enumerate}
    
\begin{figure}[h]
        \centering
        \includegraphics[width=\textwidth]{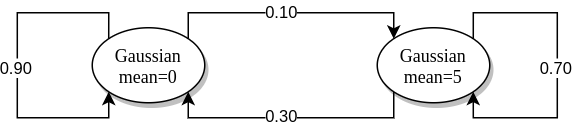}
        \caption{The markov chain diagram describes the example of process of generating time series that are bi-modal distributed and whose state is markov chain process.}
        \label{fig:markovchain}
    \end{figure}

    Tests can be generated by using one or more of these components to generate time series. In principle, the more components we mix, the harder it should be for a pipeline to learn to model it. Figure~\ref{fig:markovchain} presents a time series generator that can be used to test a forecasting pipeline's robustness to (1) Gaussian noise; (5) recurrent concept drift; (6) bi-modal distributed data.
    
    \subsection{Synthesis}
    
    Synthetic time series are split into the training set and the test set. The training set is used to train a pipeline. The test set is used to transform the input, apply transformations and predict future timesteps. 
    
    The only thing the user needs to specify is the pipeline he wishes to test. This pipeline will be tested by several time series generators three times. The only thing that remains the same between the tests is the logic necessary to generate similar time series. The scale, intercept, seasonality, trend slope, etc is not guaranteed to be the same. If a pipeline can learn this logic, and forecast values near the oracle then we consider it a good fit. A forecasting pipeline capable of learning these artificially introduced patterns amidst noise should also be able to learn them when they occur in a real-world scenario.
    
    \subsection{Evaluating performance}

    We know how to generate that data. But we want to find how to predict that data generation process, as much as possible. Given that it's an artificially controlled process, we also have access to the form time series without the noise. This provides us with absolute oracles, which can rarely be found in real-world problems. 
    
    We measure the performance of forecasting pipelines on the testing period by estimating the symmetric Mean Absolute Percentage Error (sMAPE) between the predicted values and the time series without noise (specified oracle). We chose to use sMAPE, due to its popularity in point forecasting, and due to being relative and symmetric.
    
    Forecasters should attempt to modeling to approximate the oracle line, i.e. the expected value. This is not an easy task, as tests will frequently introduce perturbations to make this task harder.
    
    \subsection{Quick start and implementation details}
    
    \textit{TimeGym} can be found on \href{https://github.com/diogoseca/TimeGym}{github.com/diogoseca/TimeGym}. We hope to contribute to open-source with the \emph{TimeGym} testing and debugging tools for forecasters. We appreciate all contributions.


    \begin{figure*}
        \centering
        \includegraphics[width=\textwidth]{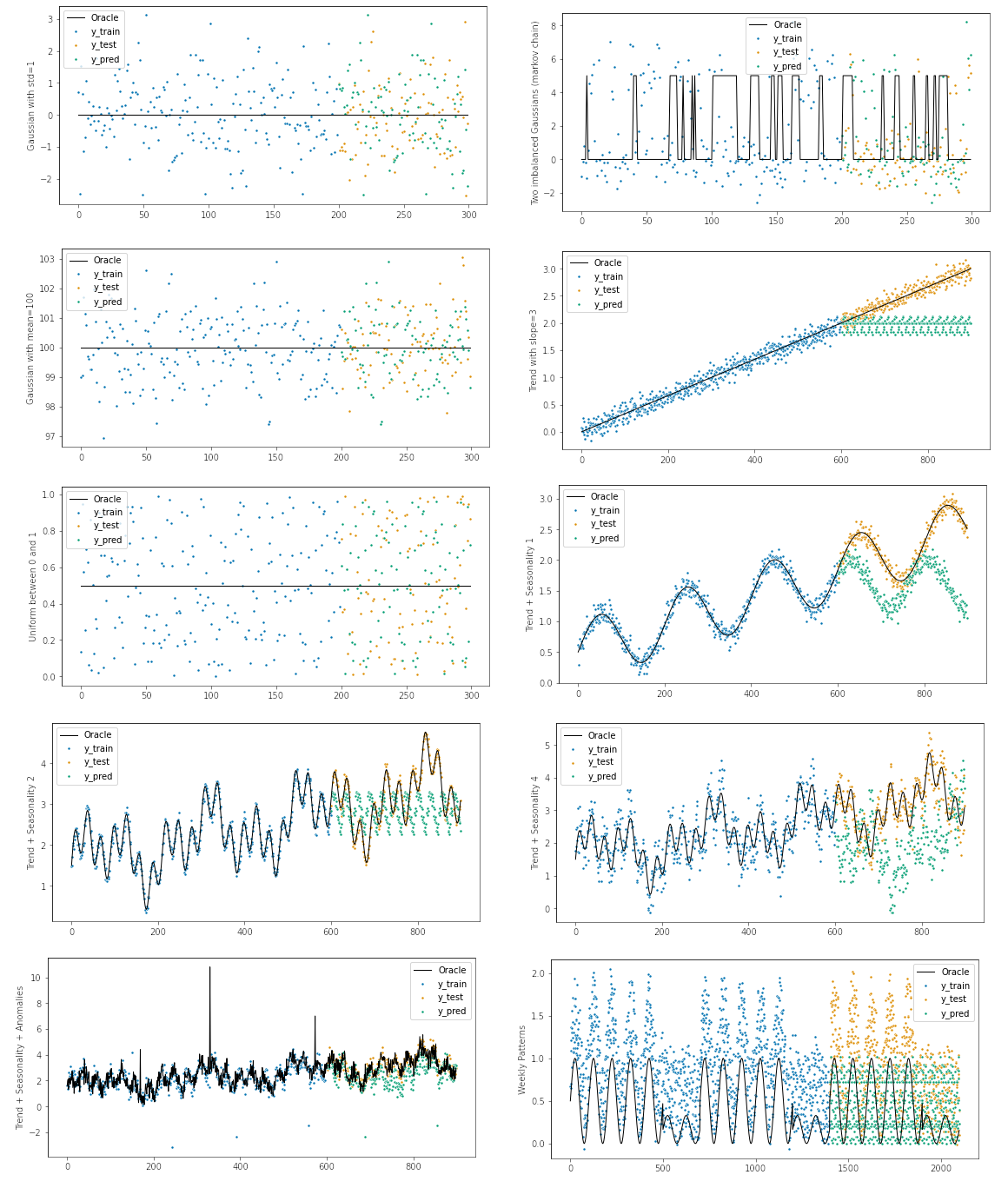}
        \caption{\textbf{K-Nearest Neighbors using a reduced recursive window of size 12.} The left axis shows the challenge name. The dark line shows the oracle, i.e. the optimal predictive values.}
        \label{fig:knn}
    \end{figure*}
    
\section{Discussion}
    
    Testing a pipeline outputs a matrix plot indicating which of the methods are sensitive to which challenges (e.g. noise, changes in trend), and which are more robust (measured by the sMAPE for the test set).

    
    
    Figure~\ref{fig:knn} shows a sample of tests applied on a forecasting pipeline that included K-Nearest Neighbors. The large deviations from the oracle line (the black line) provide evidence that the forecasting pipeline is incapable of generalizing given the training set provided. Two common problems are not having enough relevant data, and failing to learn from the data. This is not a problem for synthetic datasets, as we can freely increasing the sample size. With larger sample sizes we may also experiment with increasing modeling complexity. A great example of this modeling strategy is GPT-3 model, which achieved surprising results in text processing \cite{brown2020language}.
    
\begin{figure}[h]
        \centering
        \includegraphics[width=\textwidth]{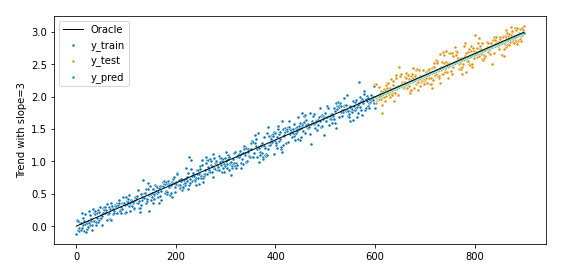}
        \caption{\textbf{AutoARIMA with a seasonal differencing period of size 12.} This pipeline seems robust to trends and noise than the KNN pipeline.}
        \label{fig:arima_trend_noise}
    \end{figure}

    One design flaw found is the incapacity to generalize the expected value of a Gaussian distribution. K-Nearest Neighbors doesn't converge to estimate the mean of the previous timesteps. Another failure in this forecasting pipeline is recognizing the trend slope. Contrast this with Figure~\ref{fig:arima_trend_noise} which shows a pipeline based on Auto-ARIMA that aptly learns the slope of a trend amidst noisy observations.

    \subsection{Meta-learning using synthetic time series}
    
    Edsger Dijkstra is quoted saying "Program testing can be used to show the presence of bugs, but never to show their absence"~\cite{dijkstra1970ewd}. Even if we come up with a forecasting pipeline that correctly models all of the tests, this model might only still be fit for predicting on our tests. Real data is still important to validate the modeling assumptions.
    
    However, the more tests a forecasting method passes, the better it should be at learning from the sample provided. If a practitioner is worried about the performance of his forecasting pipeline, it is advised to include tests based on patterns he hopes to model. If his forecasting method fails at learning these patterns in a controlled testing environment, then how can he expect it to learn anything from noisy data from the real world?
    
    Meta-learning, i.e. learning how to learn to forecast using synthetic data is, as far as we know, an unexplored topic. Automatic forecasting has touched on this research topic but as of this writing, forecasting still requires human expertise \cite{hyndman2007automatic}.

    Automated testing for forecasting will not solve issues such as data hunger, capacity brittleness, dubious input data, and fickle trust by users \cite{kolassa2020will}. However, contrary to the opinion of \cite{hyndman2018forecasting}, we show evidence that automated testing helps forecasters stay clear from using modeling procedures that result in biased learning.

\section{Conclusion}

    We present a toolbox for debugging forecasting pipelines in Python. This enables forecasters to use development processes such as Test-Driven Development, and enjoy its benefits during forecasting modeling.
    
    \textit{TimeGym} can be used at any time to uncovering flaws in the pipeline design, e.g. failing to recognize the cyclical patterns in the time series. \textit{TimeGym} can also be used as a test-driven development framework for researching designing novel forecasting methods. 
    
    We propose that future research look at how we measure objectively quantify how good are the tests in test-driven machine learning forecasting. Challenging tests use complex time series with several components that are more difficult to predict. On top of this signal, adding noise to the time series will only increase this difficulty. How much noise can we add before the time series before completely unpredictable?

\bibliographystyle{acm}
\bibliography{references} 
\end{document}